\documentclass[letterpaper, 10 pt, conference]{ieeeconf}  

\IEEEoverridecommandlockouts                              

\usepackage{float}
\usepackage{lipsum}
\usepackage{graphicx}
\usepackage{flushend}
\usepackage[hidelinks]{hyperref}
\usepackage{subcaption}
\usepackage{amsmath} 
\usepackage{amssymb}  
\usepackage{etoolbox}
\usepackage{threeparttable}
\usepackage[noadjust]{cite}
\makeatletter

\usepackage{cancel}
\usepackage{comment}
\usepackage{booktabs, dcolumn}

\newcommand\figref{Fig.~\ref}

\title{\LARGE \bf
VPE: Variational Policy Embedding for\\ Transfer Reinforcement Learning
}

\DeclareMathOperator*{\argmax}{arg\,max}

\author{
Isac Arnekvist,
Danica Kragic
and Johannes A. Stork$^{1}$
\thanks{$^{1}$Authors are with the Robotics, Perception, and Learning lab,
        Royal Institute of Technology, Sweden. {\tt\footnotesize \{isacar,jastork,dani\}@kth.se}.
        Correspondence to {\tt\footnotesize isacar@kth.se}.}
}

\begin{document}

\maketitle
\thispagestyle{empty}
\pagestyle{empty}

\begin{abstract}
Reinforcement Learning methods are capable of solving complex problems,
but resulting policies might perform poorly in environments that are even slightly different. In
robotics especially, training and deployment conditions often vary and data
collection is expensive, making retraining undesirable.
Simulation training allows for feasible training times, but on the other hand
suffers from a reality-gap when applied in real-world settings. This raises the
need of efficient adaptation of policies acting in new environments.

We consider this as a problem of transferring knowledge within a family of similar Markov
decision processes. For this purpose we assume that
Q-functions are generated by some low-dimensional latent variable. Given such a Q-function, we can
find a master policy that can adapt given different values of this latent variable. Our method
learns both the generative mapping and an approximate posterior of the latent
variables, enabling identification of policies for new tasks by searching only in
the latent space, rather than the space of all policies.
The low-dimensional space, and master policy found by our method enables policies to quickly adapt
to new environments. We demonstrate the method on both a pendulum swing-up task
in simulation, and for simulation-to-real transfer on a pushing task.

\end{abstract}

\section{Introduction}

Deep Reinforcement Learning (RL) has been successful in solving a range of
complex problems \cite{alphago,atari,levine2016end,ddpg,rl_survey}.
Unfortunately, the performance of learned policies may degrade quickly in a
slightly different test environment \cite{reality_gap}. Also, training is both
computationally intense and need considerable amounts of data, making
retraining in new environments time consuming. For most real-world settings,
such as robotics, difference in training and deployment conditions is common
and data collection is costly. This makes fast adaptation and generalization to
new environments with few interactions and little computation an important
challenge. 

To address these issues, it has been suggested to learn a single policy that
generalizes well to similar environments \cite{tobin2017domain,
antonova2017reinforcement, tajbakhsh2016convolutional}, fine-tune policies in
new environments \cite{maml, reptile, neitz, hausman2018learning}, or learn from
teacher policies \cite{one_shot_imitation, hausman2017multi, gan_imitation, linear_combination_policies}.
Despite recent progress, methods are still limited, e.g., to optimization in
large parameter spaces, restrictions in environment differences, types
of teacher policies and optimization objectives.

\begin{figure}
    \centering
    \includegraphics[width=0.49\textwidth]{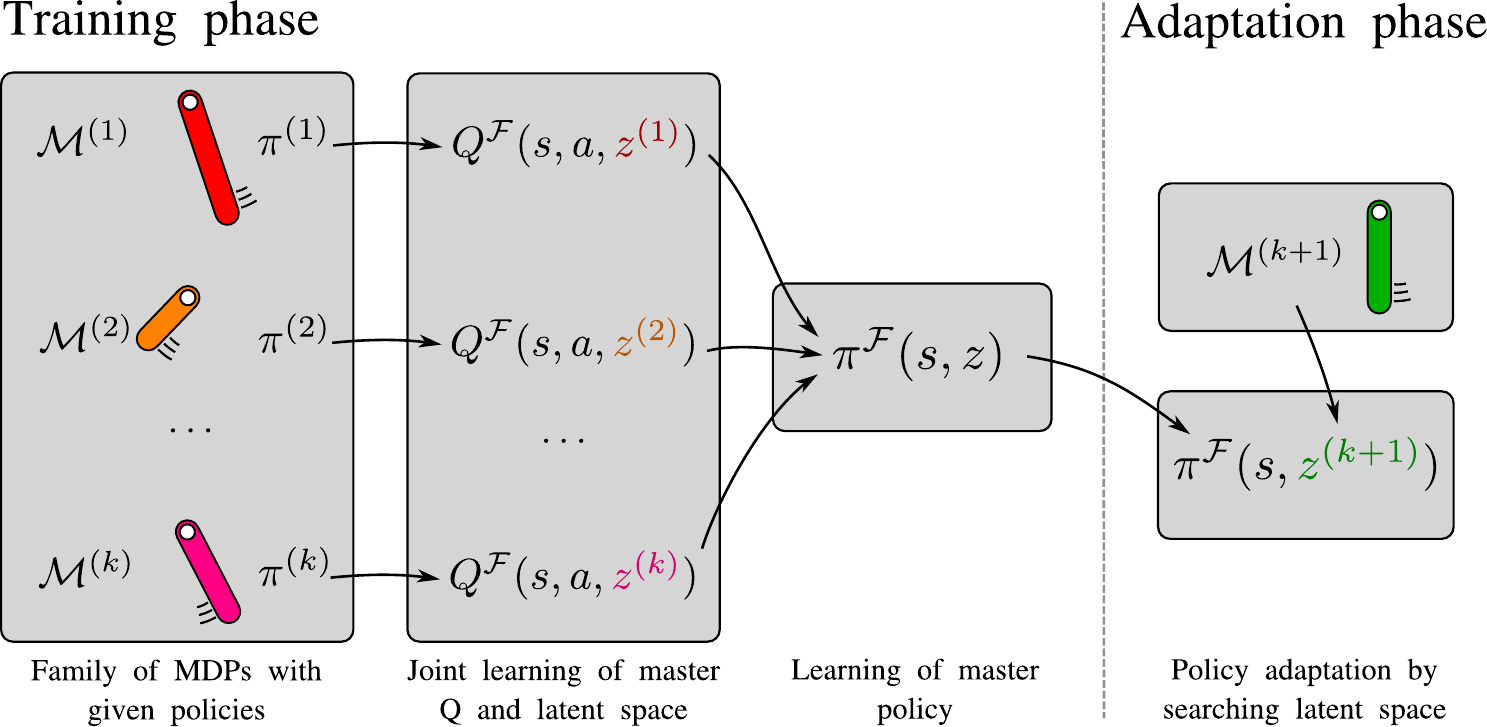}
    \caption{We are given a set of MPDs $\mathcal{M}$ along with corresponding optimal
             policies $\pi$. All MPDs share the same state and action space, but
             potentially differ in transition probabilities and reward functions. These
             differences are assumed to be generated from an unobserved variable $z$, hence
             generating a family of MDPs. By learning $z$ and how it generates Q-functions,
             we are able to transfer knowledge from teacher policies to novel MDPs within
             the same family.}
    \label{fig:mdp_family}
\end{figure}

In this work, we propose Variational Policy Embedding (VPE) for learning an
adaptable \textit{master policy} for a family of similar Markov Decision
Processes (MDPs). Instead of finding one robust policy, the master policy can
be easily adapted to new members of the family. Assuming that the family is
parameterized by some continuous latent variable, we generalize from optimal
teacher policies for a set of example members, as illustrated in
\figref{fig:mdp_family}. For example, a family of swing-up pendulum
environments could be parameterized by the mass of the pendulum and the cost to
apply a certain torque. Note here that the parameterization can influence both
transition probabilities and reward functions.

Instead of identifying the complex relationships between the latent parameter,
the family of MDPs, and their optimal policies, we use unsupervised learning to
identify a suitable embedding into a latent space, $\mathcal{Z}$. The latent
space can then be used to change the behavior of the master policy. Since
different points in the latent space encode optimal policies for the whole
family, adaptation for a new member can be carried out in $\mathcal{Z}$ instead
of searching the space of all policies. For efficient adaptation it is
therefore desirable that $\mathcal{Z}$ is low-dimensional and that its
structure is suitable for fast adaption, e.g., by having a locally smooth
injection to the space of relevant policies. To achieve this, we use
variational Bayesian methods to learn a minimum description length embedding. 

Our method is based on the following contributions:
\begin{itemize}
\item deriving an evidence lower bound for variational approximation of the latent space,
\item enabling lower bound optimization for both stochastic and deterministic teacher policies,
\item formulating policy adaption as global optimization in latent space, and
\item adapting the policy by either Bayesian optimization or stochastic gradient descent.
\end{itemize}

Our empirical evaluation shows that we can learn the master policy from a set
of teacher policies and successfully adapt it to new MDPs with only a few
optimization steps. For evaluation, learning the latent space and master policy
is carried out in synthetic domains but we also show policy adaptation in a
real-world robotic manipulation scenario.


\section{Related Work}

\textit{Transfer learning} (TL) deals with knowledge transfer from previous
learning, to lessen the need of learning tasks \textit{tabula rasa}
\cite{tl_survey}. TL have in recent years shown remarkable success in
supervised learning for vision \cite{oquab2014learning, long2015fully,
karpathy2014large}. TL is not only used for supervised learning, but is also a
relevant area of research for RL \cite{rl_transfer_survey}. In
the most general case for RL, knowledge is transferred between MDPs with
different state and action spaces. We will on the other hand restrict our
attention to the case where these are shared over similar MDPs.

\textit{Domain randomization} attempts to find a single policy that works for
all instances of a family \cite{tobin2017domain,antonova2017reinforcement}.
This is feasible in families with small variations in for example friction, but
not where good actions are necessarily different between MDPs. We will instead
consider finding a policy which can easily be adapted to new MDPs.

In \textit{meta-learning} for RL, the most recent approaches consider learning
parameter initializations that only require few gradient steps to change to a
new behavior \cite{maml,reptile}. This has been successfully demonstrated for
RL \cite{maml}, but requires second order gradients. An attempt to simplify the
method using only first order gradients was also presented for supervised
learning, but was not able to produce successful policies in the RL domain
\cite{reptile}. In addition to using second order gradients for RL, these
approaches currently update all parameters, limiting it to gradient descent
methods in high-dimensional spaces. Also, training a policy in multiple
scenarios simultaneously could be not only challenging, but impossible if
environments can not be interacted with simultaneously.

\textit{Imitation learning} could be used to learn from a set of teacher
policies trained in advance \cite{one_shot_imitation, hausman2017multi,
gan_imitation, linear_combination_policies}. For imitation learning with
supervised learning \cite{one_shot_imitation}, common loss functions can,
however, easily lead to sub-optimal actions. For example, consider approaching
a T-junction where demonstrations shows turning either left or right.
Regression with a mean squared error loss would result in a policy that drives
straight ahead. Instead of learning the mean it was proposed to learn a linear
combination of teacher policies \cite{linear_combination_policies}. Although
avoiding to learn sub-optimal means, the combination vector grows with the
number of polices. In our work, the latent space instead only needs as many dimensions
as is sufficient to distinguish environments. Another alternative to ill-behaving loss functions is to
learn the loss function and a multi-modal policy simultaneously using
generative adversarial networks (GANs) \cite{hausman2017multi, gan_imitation,
goodfellow2014generative}. GANs are however notoriously hard to train and
convergence is not guaranteed \cite{mescheder2017numerics}. We will instead
leverage well known methods for supervised learning with temporal
difference \cite{sutton1998reinforcement} and variational inference \cite{vi}.



Previous works have also considered extending Q-functions with a latent space,
allowing it to encompass multiple environments and tasks by changing the value
of the latent variable \cite{neitz,hausman2018learning}. Neitz \cite{neitz}, in
contrast to our work, only consider discrete action spaces and does not
introduce any principled way to enforce a smooth embedding and minimum
description length parameters of the latent variables. Hausman \emph{et al}.
\cite{hausman2018learning} on the other hand employs variational inference to
allow principled inference of the latent parameters. This is, however,
maximizing a trade-off between policy entropy and future rewards instead of
solely the discounted sum of future rewards. Also, Hausman \emph{et al}. lower bound
the entropy of policies, which applies to stochastic policies. In our work, we
present an alternative probabilistic formulation allowing generalization from
both stochastic and deterministic policies through optimization of the
evidence lower bound.

\section{Problem Formalization and Notation}

We consider a \textit{family} of similar MDPs, $\mathcal{F}$, where each member $\mathcal{M} \in \mathcal{F}$ is defined by a tuple,
\begin{equation*}
\mathcal{M}
\triangleq
\left( S, A, p_S, p_R, \gamma \right)
,
\end{equation*}
with (continuous) state space $S$ and action space $A$. The constant $\gamma \in \mathbb{R}^+$ denotes the discount factor and a policy is either a mapping $\pi \colon S \to A$
or a distribution over actions $\pi \colon S \times A \to [0, 1]$. If not stated otherwise, we refer to the deterministic policy.
While all members of the family share the same state space and action space, they have different transition probabilities $p_S(s_{t+1} | s_t, a_t)$ or reward distributions

$p_R(r_t | s_t, a_t)$. In all cases, subscript $t$ denotes time and both
$p_S$ and $p_R$ are stationary. In our model, the family $\mathcal{F}$ is
parameterized by some unobserved variable $Z$ from the domain $\mathcal{Z}$. For instance, the family of
inverted pendulums used in \mbox{Sec. \ref{sec:pendulum_experiment}} is parameterized
by the mass of the pendulum and the cost of applying a certain torque. 

In the policy adaption scenario we are given or can obtain $K$ optimal
policies, $\lbrace \pi^{(i)}\rbrace_{i=1}^K$, for a sample set of family
members, $\lbrace \mathcal{M}^{(i)}\rbrace_{i=1}^K \subset \mathcal{F}$. We
call these \emph{teacher policies} and \emph{teacher MDPs}. For some new MDP
$\mathcal{M}^{(K+1)} \in \mathcal{F}$, our goal is to generalize from the known
teacher policies to a new optimal policy $\pi^{(K+1)}$ for
$\mathcal{M}^{(K+1)}$ by searching a low-dimensional space and using only few
interactions with $\mathcal{M}^{(K+1)}$. To this end, we introduce one single
adaptable master policy $\pi^{\mathcal{F}} \colon S \times \mathcal{Z} \to A$ that can be
adapted to different MDPs of the family. In this formulation, adapting the policy $\pi^{\mathcal{F}}$ to $\mathcal{M}^{(K+1)}$ equals searching $\mathcal{Z}$ for the correct value $z^{(K+1)} \in \mathcal{Z}$.

\section{Variational Policy Embedding}
\label{sec:method}

In this section, we explain how we learn the master policy $\pi^{\mathcal{F}}$ for a family of similar MDPs $\mathcal{F}$, how we identify the structure of the latent space $\mathcal{Z}$, and how we search $\mathcal{Z}$ to adapt to a new and unseen member $\mathcal{M}^{(K+1)}$.
Unfortunately, directly modeling $\pi^{\mathcal{F}}$ by interpolating or combining the
teacher policies is difficult, not least because multiple optimal policies can
exist. Optimal state-action value functions (Q-functions) are on the
other hand unique for any MDP \cite[p.~50]{sutton1998reinforcement}, which
makes modeling of one single master Q-function easier. 
For this reason, we first learn a master Q-function, $Q^{\mathcal{F}} \colon S \times A \times \mathcal{Z} \to \mathbb{R}$, for the family $\mathcal{F}$ as a critic for optimizing $\pi^{\mathcal{F}}$ as an actor \cite{grondman2012survey}.


We proceed step-by-step, beginning with
jointly learning the embedding into $\mathcal{Z}$ and the function $Q^{\mathcal{F}}$ from interactions with the teacher MDPs (\mbox{Sec. \ref{sec:infer_q_z}}).
In the following step we optimize the master policy $\pi^{\mathcal{F}}$ (\mbox{Sec. \ref{sec:fitting_global_policy}}), and finally, we present two methods of
inferring $Z$ for interactions with the test MDP $\mathcal{M}^{(K+1)}$ in order to adapt $\pi^{\mathcal{F}}$ (\mbox{Sec. \ref{sec:policy_adaptation}}). Implementation details of these steps are given in \mbox{Sec. \ref{sec:implementation}}.

\subsection{Latent Space and Master Q-function}
\label{sec:infer_q_z}

\begin{figure}[]
    \centering
    \includegraphics[height=80pt]{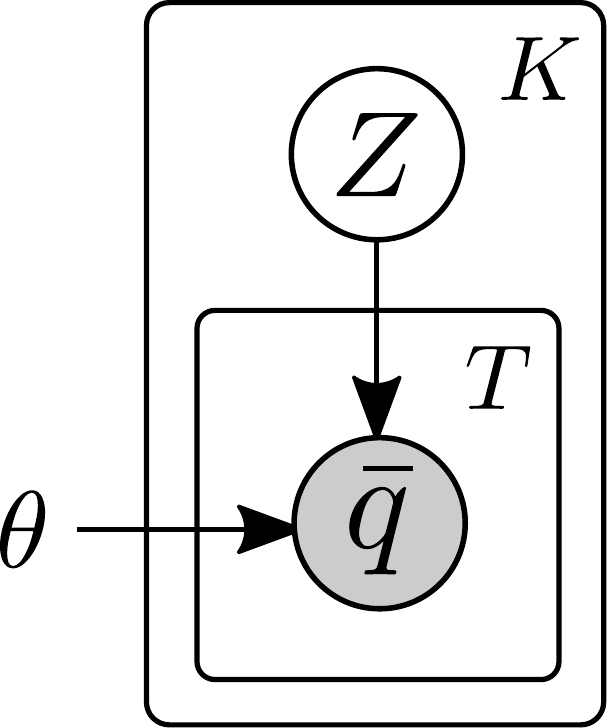}
    \caption{We consider Q-functions to be generated by a latent variable $Z$. $K$ is the number of teacher MDPs and $T$ is the number of observed transitions.}
    \label{fig:pgm}
\end{figure}

In order to learn the latent space embedding to $\mathcal{Z}$ and the master
Q-function for the family $\mathcal{F}$ we introduce a generative map from
latent space to Q-functions as seen in \figref{fig:pgm}. For this model, we
represent the master Q-function, $Q^{\mathcal{F}}$, as a neural network
with parameters $\theta$. Our learning objective is that, for every MDP
$\mathcal{M}^{(i)}$, the master Q-function matches the Q-functions that
are induced by the known teacher policies $\pi^{(i)}$,

%
%
%
\begin{align*}
Q^{(i)}_{\pi^{(i)}}(s_t, a_t)
&= 
\underset{%
p^{(i)}_S, p^{(i)}_R, \pi^{(i)}
}{\mathbb{E}}
\left[\sum_{h=0}^\infty \gamma^h r_{t+h} \mid s_{t+h}, a_{t+h}\right]
,
\end{align*}
with the learned parameters $\theta$ and latent coordinate $z^{(i)} \in \mathcal{Z}$.

However, in our setting, we only have access to teacher policies $\pi^{(i)}$
and interactions with the teacher MDPs $\mathcal{M}^{(i)}$. The induced
Q-functions $Q^{(i)}_{\pi^{(i)}}$ are not observed directly. For this reason,
we instead consider state transitions and employ the temporal difference method
\cite{sutton1998reinforcement}. An observed value for a transition $(s^{(i)}_t,
a^{(i)}_t, s^{(i)}_{t+1}, r^{(i)}_t)$ is then given as
\begin{align}\label{eq:q_targets}
\bar{q}^{(i)}_t
&
=
r_t^{(i)}
+ 
\gamma \, Q^{\mathcal{F}} \left(s_{t+1}^{(i)}, \pi^{(i)}(s_{t+1}^{(i)}), z^{(i)}\right) 
,
\end{align}
which allows us to use $\bar{q}^{(i)}_t$ as a target for $Q^{\mathcal{F}} \left(s_t^{(i)}, a_t^{(i)}, z^{(i)} \right)$ in learning.
Note that we are not necessarily restricted to deterministic teacher policies. For stochastic policies, we instead
use Monte Carlo sampled targets
\begin{align}
\bar{q}^{(i)}_t
&
=
r_t^{(i)}
+ 
\gamma \, Q^{\mathcal{F}} \left(s_{t+1}^{(i)}, a, z^{(i)}\right) 
,
\end{align}
with sampled actions $a~\sim~\pi^{(i)}(\cdot \mid s_{t+1}^{(i)})$.

\subsubsection*{Probabilistic Modeling}

We model the problem of jointly learning the embedding to $\mathcal{Z} = \mathbb{R}^d$ and the parameter $\theta$ as estimating the latent space posterior distribution. Accordingly, we define $\mathbf{Z} = \{ Z^{(i)} \}_{i=1}^K$ as the set of latent variables and $\mathbf{Q} = \{ \bar{q}^{(i)}_t : i \in 1, \dots K, t \in 1, \dots T \}$ as the set of observations as seen in \figref{fig:pgm}. The dataset $\mathcal{D}$ consists transitions $\tau = (s^{(i)}_t, a^{(i)}_t, s^{(i)}_{t+1}, r^{(i)}_t, i)$ for different MDPs. 

For the prior distribution we select a multivariate Gaussian with unit variance, $p(\mathbf{Z}) =  \prod_{i=1}^K p(Z^{(i)}) = \prod_{i=1}^K \mathcal{N}(\mathbf{0, I})$, and approximate the posterior distribution $p(\mathbf{Z} | \mathcal{D})$,
\begin{equation}
\label{eq:approximation}
p(\mathbf{Z} | \mathcal{D})
\approx
q(\mathbf{Z}) 
= 
\prod_{i=1}^K q_i(Z^{(i)}) 
= 
\prod_{i=1}^K \prod_{j=1}^d \mathcal{N}(\mu^{(i)}_j, \sigma_j^2)
,
\end{equation}
with one mean vector, $\mu^{(i)} \in \mathbb{R}^d$, for each teacher MDP and a shared vector of variances for each dimension, $\sigma  \in \mathbb{R}^d$. The likelihood, $p(\mathcal{D} | \mathbf{Z}, \theta)$, is defined as a product over all transitions in the dataset,
\begin{align}
p(\mathcal{D} | \mathbf{Z}, \theta)
=
\prod_{\tau \in \mathcal{D}} 
p( \bar{q}^{(i)}_t | s^{(i)}_t, a^{(i)}_t, z^{(i)}, \theta) 
.
\end{align}
For each transition we model the likelihood with a Gaussian,
\begin{align}
p(\bar{q}^{(i)}_t | s^{(i)}_t, a^{(i)}_t, z^{(i)}, \theta) 
= 
\mathcal{N}
(
Q^{\mathcal{F}}(s_t^{(i)}, a_t^{(i)}, z^{(i)}), \frac{1}{\lambda}
)
,
\end{align}
where $\lambda \in \mathbb{R}$ is a small, a-priori chosen constant.

\subsubsection*{Probabilistic Inference}

We infer the parameters of the approximated posterior $\{ \mu^{(i)} \}_{i=1}^K$ and $\sigma$  together with the neural network parameters $\theta$ by maximizing the evidence lower bound (ELBO),
\begin{align}
\mathcal{L}_{\text{ELBO}}(\theta, \mu, \sigma) 
={}
\sum_{\tau \in \mathcal{D}}
\Big(
    &\underset{
    \substack{
    z \sim q_i}
    }{\mathbb{E}}
    \left[
    \log p(\bar{q}^{(i)}_t | s^{(i)}_t, a^{(i)}_t, z, \theta)
    \right] +
    \notag
    \\
    &
    D_{\text{KL}}\left( q_i(Z^{(i)})~\|~p(Z^{(i)}) \right) \Big)
\end{align}
%
However, by ignoring constants, we can equivalently maximize the following objective:
\begin{align}
	\mathcal{L}(\theta, \mu, \sigma) =
    \sum_{%
    \substack{%
    \tau \in \mathcal{D}\\
    }} 
    \bigg( &-\frac{\lambda}{n}\sum_{z\sim q_i}\left( q_t^{(i)} - Q^{\mathcal{F}} (s_t^{(i)}, a_t^{(i)}, z)\right)^2  + \nonumber \\
    &\sum_{j=1}^d \left( \sigma_d^2 + \mu_d^{(i)2} - \ln \sigma_d^2 \right) \bigg)\label{eq:elbo_prop_loss}
,
\end{align}
where we employ Monte Carlo integration by randomly drawing $n$ samples $z~\sim~q_i$ in \mbox{Eq. \eqref{eq:elbo_prop_loss}}.

Optimization of the objective in \mbox{Eq. \eqref{eq:elbo_prop_loss}} over the parameters $\mu$, $\sigma$, and $\theta$ serves two purposes. First, it finds parameters of $Q^{\mathcal{F}}$  that are compliant with the latent space embedding to $\mathcal{Z}$. Second, the optimization is equivalent to minimization of the divergence between the true posterior and the approximate posterior,
\begin{equation}
D_{\text{KL}}\left( q(\mathbf{Z})~\|~p(\mathbf{Z}|\mathcal{D})\right)
.
\end{equation}
This means that we are approximating the true posterior in terms of
KL-divergence. As a key consequence, the embedding is compressed in terms of information theoretic description length \cite{hinton1993keeping, graves2011practical}. In our evaluation in \mbox{Sec. \ref{sec:results}}, we accordingly observe that dimensions of the latent space $\mathcal{Z}$ that are not needed to model $Q^{\mathcal{F}}$ fall back to the prior.


\subsection{Learning the Master Policy}
\label{sec:fitting_global_policy}

After having learned the latent space embedding and the master Q-function for the family $\mathcal{F}$ in \mbox{Sec. \ref{sec:infer_q_z}}, we now have to learn the master policy $\pi^{\mathcal{F}}$ that, additionally to states, takes latent coordinates as input. For this, we model the master policy as a neural network with parameters $\phi$. We identify the policy parameters $\phi$ by maximizing the master Q-function similar to actor-critic learning \cite{grondman2012survey},
\begin{equation}
\label{eq:policy_maximization}
\argmax_\phi 
\underset{
\substack{
\tau \in \mathcal{D}
\\  
z \sim q_i}
}{\mathbb{E}}
\left[ 
Q^{\mathcal{F}}
\left(s_t^{(i)}, \pi^{\mathcal{F}}(s_t^{(i)}, z), z\right) 
\right]
.
\end{equation}
In practice this expectation can be computed with the training dataset $\mathcal{D}$ from in \mbox{Sec. \ref{sec:infer_q_z}}.

\subsection{Policy Adaptation in Latent Space}
\label{sec:policy_adaptation}

We propose two distinct methods of adapting $\pi^{\mathcal{F}}$ to a given new MDP $\mathcal{M}^{(K+1)}$. Both methods infer the latent coordinates $z^{(K+1)}$ but the methods differ in the optimization objective and the way they interact with the new MDP. Practical details are given in Sec.~\ref{sec:implementation}.

\paragraph*{ELBO Maximization}

In this method we search for the correct latent space coordinate by maximizing the bound from Eq.~\eqref{eq:elbo_prop_loss} for a new coordinate $\mu^{(K+1)}$. For this, we collect a dataset of transitions in the new MDP. Different from the procedure for learning the latent space embedding and the master Q-function, we keep all parameters but $\mu^{(K+1)}$ fixed.

\paragraph*{Bayesian Optimization}

In this method, we employ Bayesian optimization (BO) \cite{bo_orig} for global search in $\mathcal{Z}$. To this end, we iteratively interact with the new MDP and collect rollouts with the policy for the current value of $z^{(i)}$. The cost function is defined as rollout performance in the new MDP.
To minimize the search space we only consider latent dimensions with large enough
mean signal-to-noise ratio (SNR),
\begin{equation}
\mathrm{SNR}_d = \frac{1}{K\sigma_d}\sum_{i=1}^K |\mu_d^{(i)}|
.
\end{equation}
Latent dimensions that are distributed close to the prior, hence not providing information,
have signal-to-noise ratio close to zero, and can be disregarded.

\section{Implementation Details}
\label{sec:implementation}

In this section, we provide details of how our method described in Sec.\ref{sec:method} is implemented for the experiment in Sec.\ref{sec:experiments}. The sections are ordered according to the same structure as in Sec.\ref{sec:method}.

\subsection{Latent Space and Master Q-function}
\label{sec:adaptive_q_learning}

For each experiment, we sampled a dataset $\mathcal{D}$ of one million state transitions. Sampling was accomplished by performing $\varepsilon$-greedy rollouts in the teacher MPDs with the corresponding teacher policies. The parameter $\varepsilon$ was set to $0.5$ but we note that this might not be ideal in general. A small validation set was also collected, and model parameters with the best ELBO on this set was kept for policy fitting.

The master Q-function $Q^{\mathcal{F}}$ was modeled by a 10-layer residual network with 400 ReLU units in each layer and the latent space $Z$ had 8 dimensions. Target and tracking networks were used for both the parameters $\theta$ and the latent space parameters \cite{atari}.

Training was run for one million gradient descent updates, with each mini-batch containing 32 data points. The KL-divergence term in the loss was linearly increased from zero the first 50 thousand iterations \cite{vi_warmup}. Target values were generated according to \mbox{Eq. \eqref{eq:q_targets}} and were normalized using Pop-Art \cite{popart}. Samples from the approximate posterior, state, and action variables were concatenated as a single input to the master $Q$-function. The state and action space inputs were normalized using Welford's algorithm \cite{welford}. The observation noise parameter $\lambda$ was set implicitly by weighting the likelihood term by $10.0$, and the KL-divergence term by $0.001$. Adam \cite{adam} was used as optimizer.

\subsection{Learning the Master Policy}
\label{sec:policy_fitting}

The same dataset as described in \ref{sec:adaptive_q_learning} was used to fit
the adaptable policy $\pi^{\mathcal{F}}$. The neural network architecture was
identical to the one for $Q^{\mathcal{F}}$ with the only difference being the
input and output dimensions. Master Q-function and latent space parameters
were kept fixed while optimizing \mbox{Eq. \eqref{eq:policy_maximization}}. Training
was run for 2 million optimization steps, with a batch size of 128. Adam
\cite{adam} was also used for the actor, with the addition of a weight decay of
$0.01$. After every 100 gradient updates, an estimate of the mean return was
attained by executing rollouts with the policy. The parameters associated with
the best return are used for $\pi^{\mathcal{F}}$.

\subsection{Policy Adaption in Latent Space}

To find latent space coordinates for a new MDP $\mathcal{M}^{(k+1)}$, we
searched over possible assignments of $z^{(k+1)}$ while leaving $\phi$
unchanged. We evaluated both \textit{ELBO Maximization} and
\textit{Bayesian Optimization} (BO). For \textit{ELBO Maximization} we used
stochastic gradient descent by repeating the procedure of
Sec.~\ref{sec:policy_fitting}, \mbox{Eq. \eqref{eq:elbo_prop_loss}}, with a
dataset consisting of $16000$ state transitions in the new MDP. For evaluation,
$z^{(k+1)}$ was deterministic by setting it to the mean $\mu^{(k+1)}$.

BO  \cite{bo_orig} builds a Gaussian process (GP) posterior of the cost function. As cost
function we defined the rollout return given assignments of $z^{(K+1)}$ used the Mat\'ern kernel. As acquisition function, the upper
confidence bound of the posterior was used. For the exact details, we refer to
the framework of \cite{bayes_opt}. We used only default values as hyperparameters.

\section{Experiments}
\label{sec:experiments}

In the following, we describe two experiments to demonstrate the proposed
method. We start with a simpler family of swing-up pendulum MDPs
in simulation. We then proceed with a more challenging pushing problem,
where we perform policy adaptation on a real-world robotic system.

\subsection{Pendulum Swing-up}
\label{sec:pendulum_experiment}

We start with the classic control problem of swinging up an inverted pendulum.
Although not being a notoriously hard problem, parameterizing both the state
transition distribution and the reward distribution is intuitive by changing
the mass of the pendulum, and the relative cost of actuating the joint. With
this parameterization, we can also make sure that optimal policies for some MDPs
can not successfully get the pendulum to upright in other MDPs. This problem
also have at least two optimal policies in the state where the pendulum is at
rest, pointing down.

We modify the Open AI environment \textit{Pendulum}
\cite{openaigym} for our first family of MDPs. Let the pendulum angle from
upright be denoted $\psi$. The action is defined as the acceleration of this
angle, $\ddot{\psi}$. We choose two parameters for family generation, one
governing the transition dynamics, and one for the reward function. The
dynamics is altered by sampling the mass of the pendulum uniformly in
$\left[0.4, 1.2\right]$. The reward function is altered by a parameter $\kappa \in
\left[ 0.0, 2.0\right]$ as follows:
\begin{equation*}
    r_{\kappa^{(i)}}(\psi, \dot{\psi}, \ddot{\psi}) = -\left(\psi^2 + 0.1 \dot{\psi}^2 + \kappa^{(i)} \ddot{\psi}^2\right)
\end{equation*}

Before training, 40 teacher MDPs were sampled followed by training of teacher
policies for each of the MDPs. Teacher policies were constructed by
discretization of the state and action space and performing value iteration
with a tabular approximation of the value function \cite{value_iteration}. The
resulting policies turned out to be sub-optimal, but they all get the pendulum
to an upright, stable, position. Policies can however not successfully be used
in other MDPs within the family, as expected. In addition to the 40 teacher
MDPs/policies, a set of 4 MPDs and policies were added as a test set for policy
adaptation. The discount factor $\gamma$ was set to $0.99$. Further details
regarding teacher policy training can be found in Appendix
\ref{sec:pendulum_value_iteration_details}.

\subsection{Non-prehensile Manipulation}

\begin{figure}
	\includegraphics[width=0.49\textwidth]{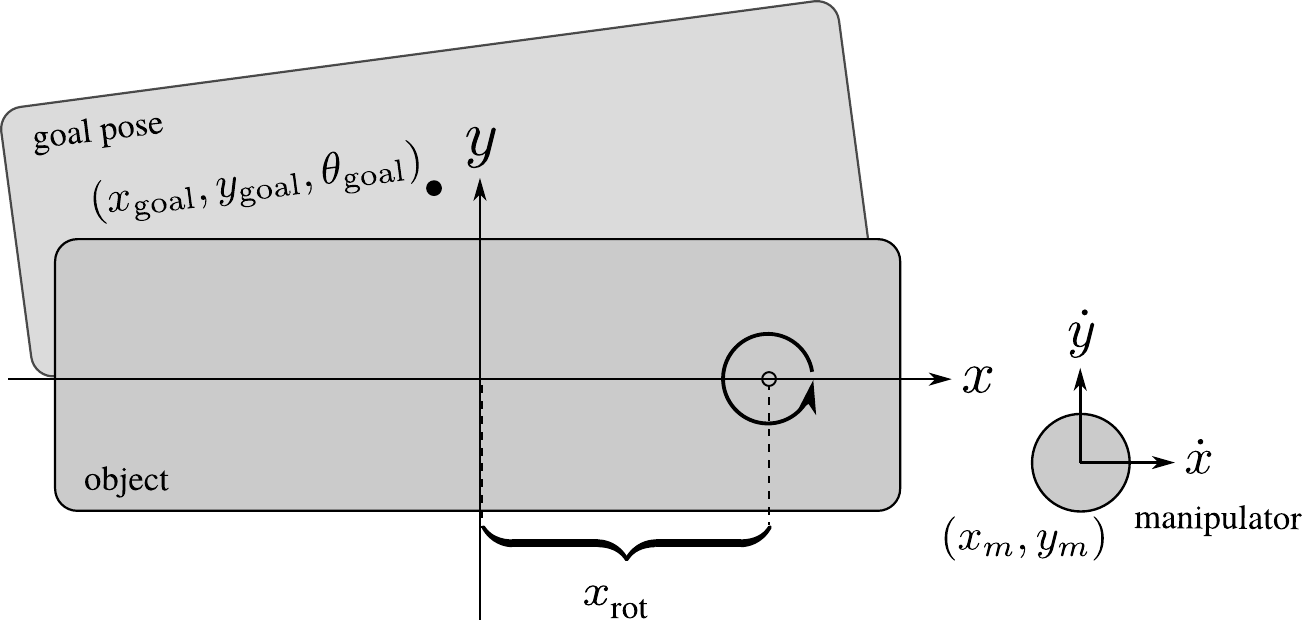}
    \caption{The second family of MDPs is based on a pushing problem.
             The aim is to control the velocities of the manipulator
             such that the object is aligned with the goal pose.
             The generation of different MDPs is done by offsetting
             the rotational joint as shown by $x_\mathrm{rot}$.}
    \label{fig:pushing_state}
\end{figure}

To illustrate the method further, we consider a more challenging problem by
training the global Q-function and policy in simulation, and then perform
evaluation by adapting to a new MDP given by a real-world robotic system. The idea of this
experiment is to increase the dimensionality of the problem, and to see if we
can accomplish transfer learning from simulation to the real setting. The
task is to push a box with varying dynamics to a fixed goal pose.  The box that
is pushed in the real-world system has a weight placed inside, off-center,
serving as the unknown latent variable. By changing the position of the weight,
the transition distribution drastically changes, and hence also the behavior of
a successful policy. In simulation, the placement of the weight is accomplished
in by offsetting the rotational axis of the box. The problem is illustrated in
\figref{fig:pushing_state}.
The state is denoted by (all in object frame):
\begin{equation*}
	s = (x_\mathrm{m}, y_\mathrm{m}, \dot{x}_\mathrm{m}, \dot{y}_\mathrm{m}, x_\mathrm{goal}, y_\mathrm{goal}, \theta_\mathrm{goal})
\end{equation*}
Here, $m$ refers to the manipulator state.

\begin{figure}
	\centering
	{%
	\setlength{\fboxsep}{0pt}%
	\fbox{\includegraphics[width=0.4\textwidth]{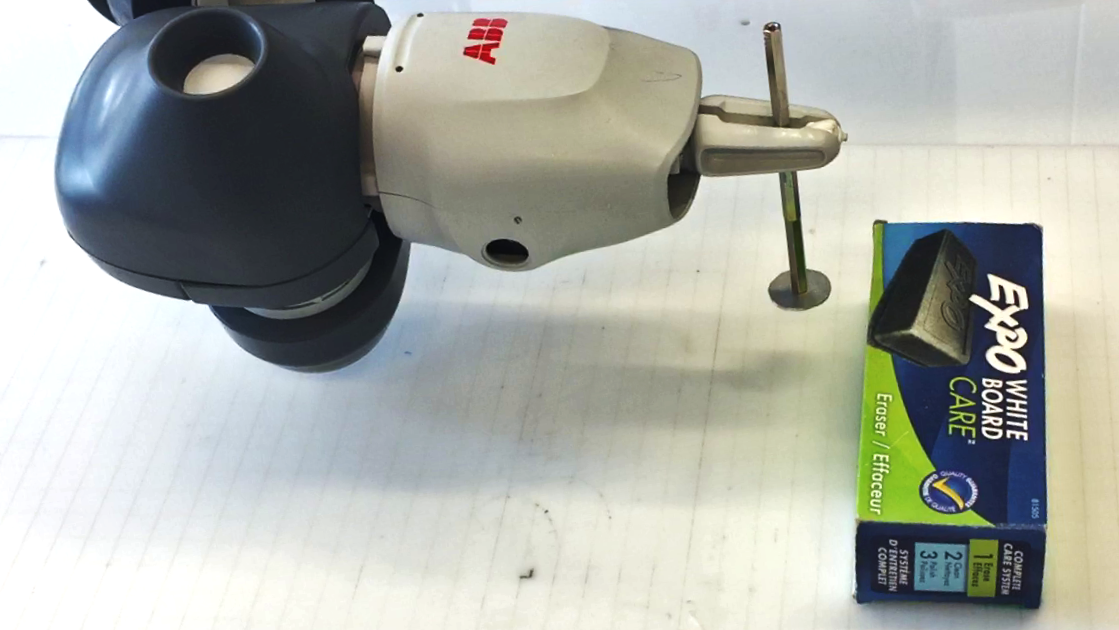}}
	}%
    \caption{For pushing, a master policy is first trained in simulation and then
             adapted on a real-world setup. We use an ABB YuMi robot to push a box using
             a planar cartesian controller.  The box has a weight inside, placed
             off-center, to make the dynamics challenging.}
    \label{fig:real_world_setup}
\end{figure}

For the teacher policies, the reward function $r$ was heavily shaped to
make training possible ($s'$ denotes successor state):
\begin{align*}
	r(s, s') = &g(s') - g(s) + h(s')\\
	g(s) = &-c_1 \cdot \|(x_\mathrm{goal}, y_\mathrm{goal}, c_2\theta'_\mathrm{goal})\|_2\\
	     &-c_3 \cdot \|(x_\mathrm{m}, y_\mathrm{m})\|_2\\
	h(s) &= c_3 \cdot \exp\left(-c_4\|(x_\mathrm{goal}, y_\mathrm{goal}, c_2\theta_\mathrm{goal})\|_2\right)
\end{align*}
The parameters above was set to: $c_1=100$, $c_2=0.1$, $c_3=10$, and $c_4=32$.
For adaptation on the real robot, this was changed to a simpler
reward based solely on the final distance to the goal after one rollout.
For simulation, we use the MuJoCo physics simulator \cite{mujoco}.
The real-world setup is shown in \figref{fig:real_world_setup}.
The offset of the rotational axis is in simulation sampled
uniformly in $[-0.05, 0.05]$, where $\pm 0.08$ corresponds to the outer edges of the box.
Also for these MDPs, the discount factor was set to $\gamma = 0.99$.

The training set policies are found by using deterministic policy
gradients (DPG) \cite{ddpg}.
Further information about learning these policies is described in
Appendix \ref{sec:pushing_teacher_policy_details}. Implementation
details regarding the robotic setup can be found in Appendix
\ref{sec:robotic_setup}.

\subsection{Policy Adaptation}

\subsubsection{Pendulum Swing-up}

As an objective function, for a given assignment of $z^{(K+1)}$,
we calculated the average return of 4 rollouts with 200 steps each.
The seed was set to zero before each call to the objective function.
The GP was fitted first after 5 initial samples, then followed
by 15 additional optimization steps. In terms of state transitions,
this totals $4 \cdot 20 \cdot 200 = 16000$, same as for the gradient descent
described above.

\subsubsection{Non-prehensile Manipulation}

For practical reasons on the real system, the objective function was defined from the final
state of one rollout (in object frame):
\begin{equation}\label{eq:yumi_return}
	f(x_\mathrm{goal}, y_\mathrm{goal}, \theta_\mathrm{goal}) = -10 \cdot \|(x_\mathrm{goal}, y_\mathrm{goal}, \frac{\theta_\mathrm{goal}}{10})\|_2
\end{equation}
This becomes zero when the goal pose is identical to the object pose, otherwise negative. The GP
was fitted after 8 rollouts of random $z^{(K+1)}$ assignments, followed by 20 additional rollouts in the optimization procedure.

\section{RESULTS}
\label{sec:results}

To demonstrate our method, we will in this section present qualitative and
quantitative results regarding both the found embedding, and the performance of
the adapted master policy. For the latent space, we want to know whether
we see compression of information to a few dimensions, according to signal-to-noise
ratio (SNR). We also want to see whether the learned embedding is comparable to the true
environment parameters.

\subsection{Pendulum Swing-up}

\begin{figure}
	\includegraphics[width=0.24\textwidth]{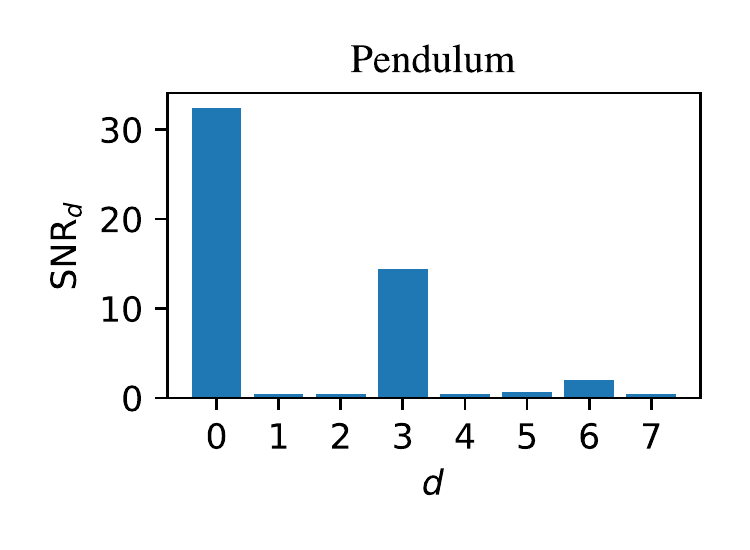}
	\includegraphics[width=0.24\textwidth]{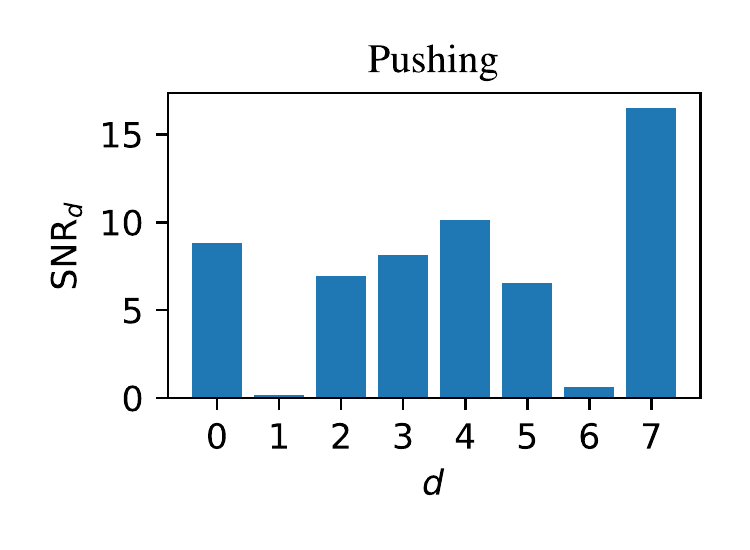}
    \caption{Mean signal-to-noise ratios (SNR) of found latent space parameters. In
             both environments, we used the two dimensions with highest SNR for
             policy adaptation. For the pendulum, dimensions 0 and 3 encode
             mass and torque cost $\kappa$. For the pushing task, dimension 7 corresponds
             to the rotational offset $x_\text{rot}$.}
    \label{fig:snr}
\end{figure}

\begin{figure}
    \centering
	\hspace*{\fill}
	\includegraphics[height=90pt]{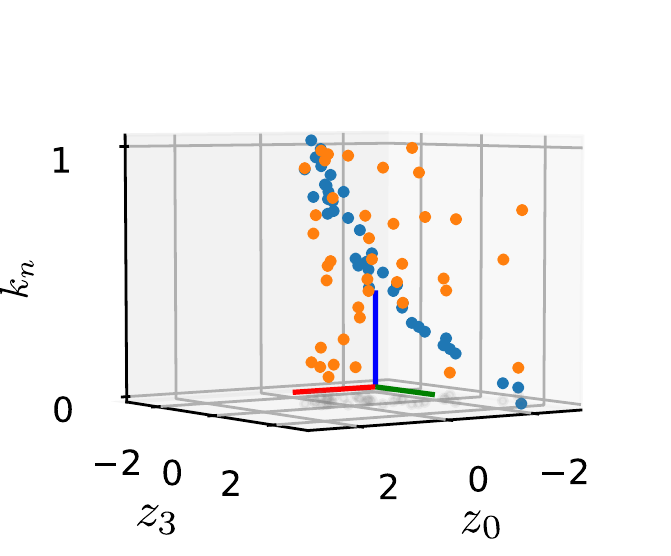}
	\hspace*{\fill}
	\includegraphics[height=90pt]{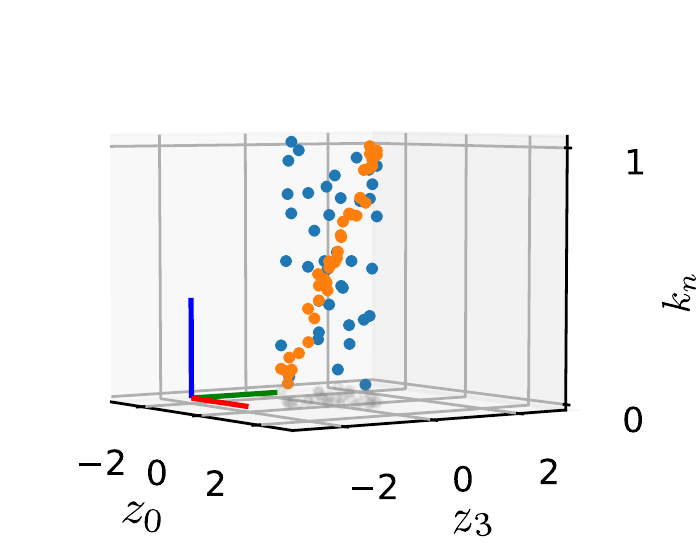}
	\includegraphics[height=110pt]{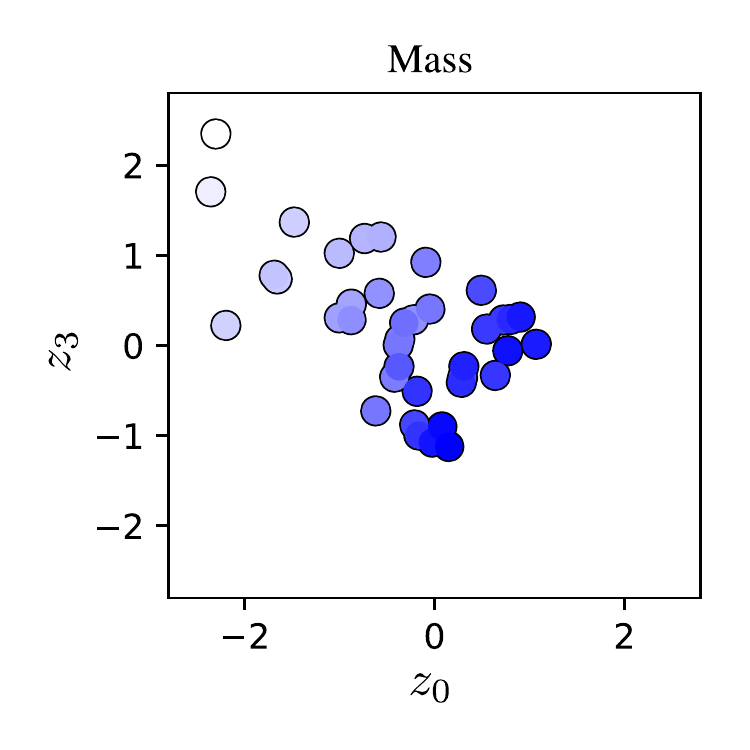}
	\includegraphics[height=110pt]{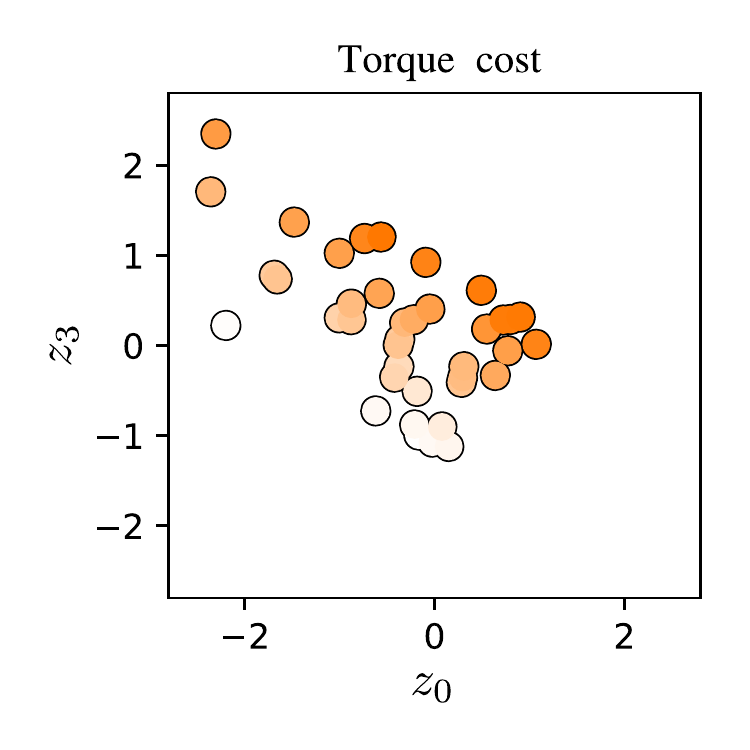}
	\hspace*{\fill}
    \caption{Found latent space parameters $z_0$ and $z_3$ plotted against pendulum
             mass (blue) and action penalty coefficient (orange). Mass
             and torque cost are normalized to be in $[0, 1]$ for illustrative purposes. The plots
             are rotated exactly $90^\circ$ in the $(z_0, z_3)$-plane to illustrate
             that these features are orthogonal, even if $z_0$ and $z_3$
             are correlated. Bottom plots are showing the same latent space
             dimensions, but color coded according to the mass and torque cost parameters.
             Best seen in color.
             }
             \label{fig:pendulum_latent}
\end{figure}

After joint learning of the master Q-function and the embedding, we
saw two peaks in the latent space, in terms of SNR.
This is shown in \figref{fig:snr}, dimensions 0 and 3.
To qualitatively assess these dimensions, we plot these dimensions
against mass and torque cost $\kappa$, shown in \figref{fig:pendulum_latent}.
The two parameters are clearly encoded by two orthogonal planes in the latent space.
The remaining six dimensions fell back to the prior and was ignored in the adaptation phase.

For policy adaptation, the results can be seen in \mbox{Table
\ref{table:pendulum_adaptation_results}}. Comparison was made between average
policies, teacher polices, SGD-trained policies, and BO-trained policies.  The
average policy value is calculated by repeated draws of $z$ from the prior,
performing a rollouts, and averaging the return. A thousand rollouts were
performed to calculate the average return and standard error. The suboptimal
teacher policies were outperformed by the global policy after adaptation with
BO. Stochastic gradient descent produced policies that in some cases were
worse than policies drawn randomly from the prior, and never outperforming the
policies found by BO.

\begin{table}
    \caption{Comparison of sampled returns in pendulum test environments}
    \label{table:pendulum_adaptation_results}
	\centering
    \scalebox{0.93}{
      \begin{tabular}{ c c c c c }
          \toprule
          \# & Average policy & Teacher policy & SGD & 20-step BO \\
          \midrule
          1 & $-134.6\pm3.0$ &  $ -97.2\pm1.6$ & $-105.4\pm1.6$ & $\mathbf{ -90.0\pm1.5}$   \\
          2 & $-161.7\pm4.2$ &  $-152.5\pm2.8$ & $-175.6\pm2.8$ & $\mathbf{-121.5\pm2.2}$   \\
          3 & $-131.0\pm3.3$ &  $-117.0\pm1.9$ & $-113.6\pm1.8$ & $\mathbf{-103.0\pm1.7}$   \\
          3 & $-188.9\pm5.0$ &  $-181.5\pm3.0$ & $-274.2\pm5.1$ & $\mathbf{-136.2\pm2.6}$   \\
         \bottomrule
      \end{tabular}
    }
\end{table}

\subsection{Non-prehensile Manipulation}

\begin{figure}[H]
	\centering
	\includegraphics[width=0.25\textwidth]{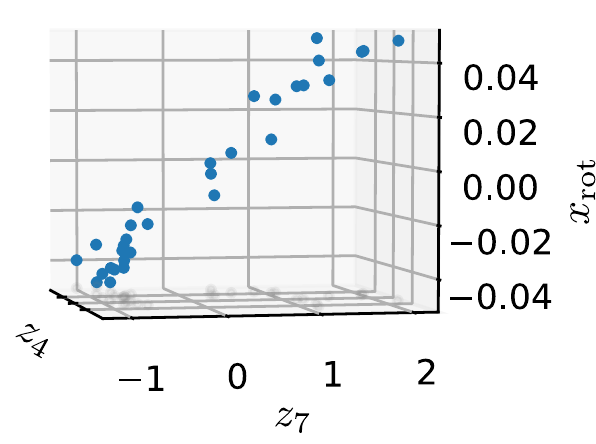}
	\includegraphics[width=0.22\textwidth]{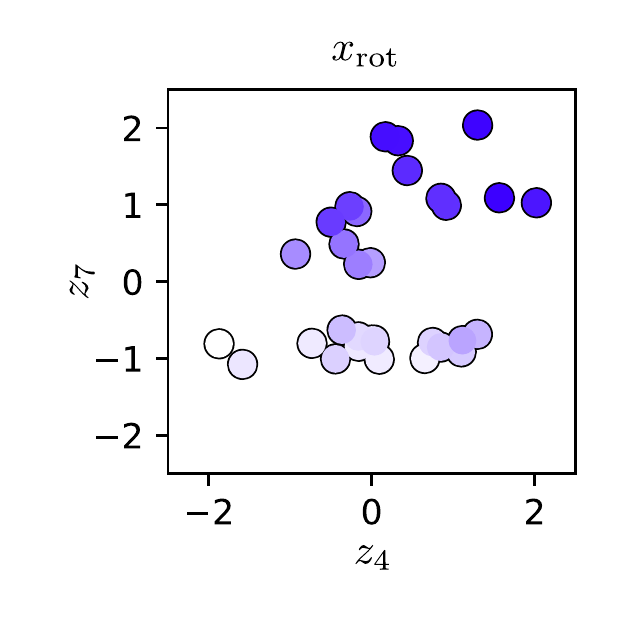}
    \caption{The parameter $x_\mathrm{rot}$ mainly encoded
             in the latent dimension $z_7$.}
    \label{fig:rotation_latent}
\end{figure}

\begin{figure}
	\centering
	\includegraphics[height=130pt]{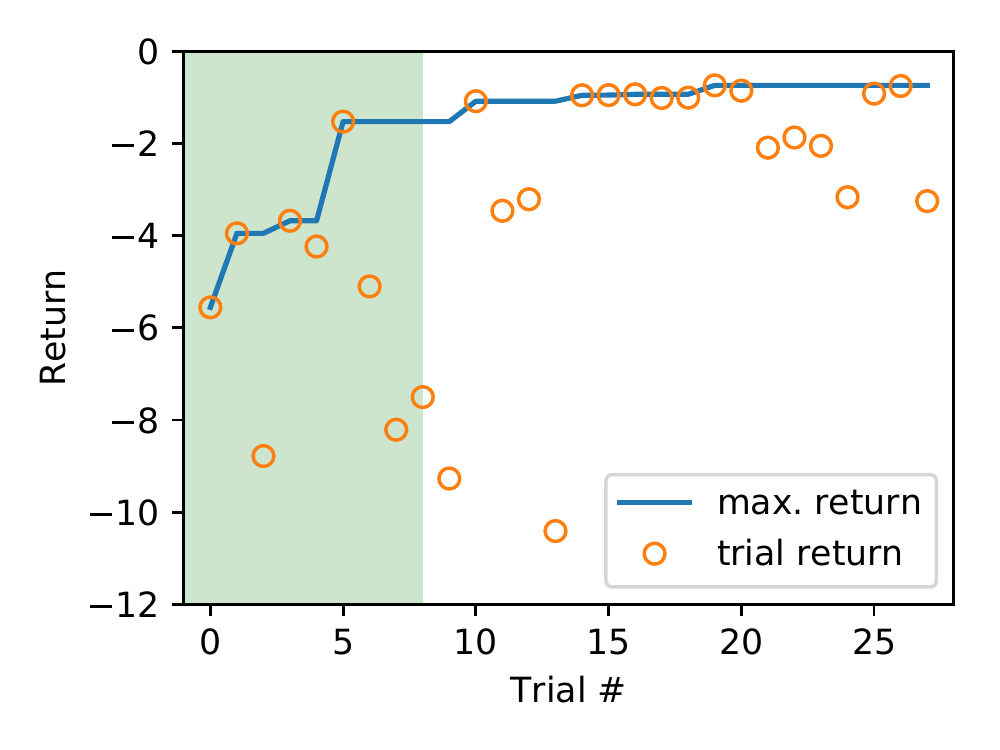}
    \caption{Bayesian optimization on the real-world pushing task.
    		 The green region represent the first random samples of
             $\mu^{(K+1)}$ before the GP is first fitted.
             Returns were calculated from the final state in each rollout
             from \mbox{Eq. \eqref{eq:yumi_return}}.}
    \label{fig:pushing_adaptation}
\end{figure}

SNR for the learned embedding is shown in \figref{fig:snr}. The two dimensions
with the highest SNR are plotted against the rotational
offset $x_\text{rot}$ in \figref{fig:rotation_latent}. Two dimensions
equals the prior, and the dimension with the highest SNR clearly encodes
the parameter $x_\text{rot}$. The other dimensions could not be
associated with $x_\text{rot}$. Also here, we used the two dimensions with
highest SNR for optimization with BO on the real robot. Results of the BO
procedure are shown in \figref{fig:pushing_adaptation}. The BO procedure, and
demonstrations of the final policy can be seen here:
\url{https://youtu.be/OMR7hHNSEKM}.

\vspace*{-3pt}
\section{Discussion}
\label{sec:discussion}

The results show that the method is able to adjust well to new
pendulum and pushing MDPs within few optimization steps using BO.
On the real robot, this corresponds
to successful policies within five to ten minutes, including both rollouts and GP optimization.
This shows that we are indeed able to learn and adapt the master policy
to new MDPs.

Regarding the latent space, our results show clear reconstructions of
environment parameters in the latent space. These parameters are encoded in the
dimensions with the highest SNR, giving us a method of determining importance
of latent dimensions and select a low-dimensional space for policy
optimization. For the pushing task, dimensions on average had higher SNR than
for the pendulum environments even though the true environment parameters were
fewer. Note, however, that the latent space encodes only environment
differences when policies are optimal. When policies are sub-optimal, the
Q-function is no longer unique, and the latent space possibly encodes both
environment parameters and policy differences.

Possible constraints of this method are increased dimensionality and MDP
complexity. Since a Q-function contains all expected returns given any action
in any state, adding on top of this the ability to interpolate between
Q-functions is clearly a challenging task.

\vspace*{-3pt}
\section{Conclusion}

To enable efficient transfer to new MDPs, we have considered a generative model
where latent parameters generate Q-functions. For this, we derived an evidence
lower bound for tractable inference of latent space parameters. This allows
transfer from stochastic and deterministic teacher policies to novel MDPs.
Lower bound optimization compresses the description length of the approximate
posterior, which we confirmed in our experiments where latent variables fall
closer to the prior. This allows us to select a small subspace suitable for
global optimization strategies. We further demonstrated empirically, both in
the synthetic domain and for simulator-to-real transfer, that we can adapt the
master policy efficiently to new MPDs.

\vspace*{-3pt}
\section{Future work}
\label{sec:future_work}
In continuation of this research we plan to more closely investigate the
consequences of teacher policy sub-optimality, consider on-line adaption
scenarios where the MDP changes over time, and explore more complex control
tasks.

\vspace*{-2pt}
\section*{Acknowledgments}
This work was funded by the Swedish Fund for Strategic Research (SSF) through
the project Factories of the Future (FACT).
The authors would also like to thank Carlo Rapisarda and Robert Krug for
help with the real-robot experiments.

\vspace*{-3pt}
\appendix
\subsection{Pendulum value iteration details}
\label{sec:pendulum_value_iteration_details}
The state was tile encoded by discretizing each dimension into both
71 and 93 bins.
The action dimension was discretized into 101 bins. The odd
number of bins is due to exclusion of the middle value, $0$,
that happens if you are dividing into an even amount.
Value iteration was run for each environment for 4 hours.

\subsection{Pushing teacher policy details}
\label{sec:pushing_teacher_policy_details}
Deterministic policy gradients (DPG) were used to construct the teacher
policies \cite{ddpg}. Parameter space noise for exploration was used,
and the architecture of the actor and critic proposed along with that
method was used \cite{parameter_space_noise}. Online normalization of states and
actions was done using Welford's algorithm \cite{welford}. Target values
were normalized using Pop-Art \cite{popart}.

\subsection{Robotic setup}
\label{sec:robotic_setup}
An ABB YuMi robot was used for the pushing experiments using
a planar cartesian controller. The object was tracked using SimTrack
\cite{simtrack}. Velocities were set
slow enough for the box to behave quasistatically.


\bibliographystyle{IEEEtran.bst} 
\bibliography{IEEEabrv,./bibliography.bib}

\end{document}